\documentclass[conference]{IEEEtran}
\IEEEoverridecommandlockouts
\usepackage{cite}
\usepackage{amsmath,amssymb,amsfonts}
\usepackage{algorithmic}
\usepackage{graphicx}
\usepackage{textcomp}
\usepackage{xcolor}
\usepackage{booktabs}
\usepackage{tabularx}
\usepackage{adjustbox}
\usepackage{makecell}
\usepackage{svg}
\usepackage{multirow}
\usepackage{multicol}
\usepackage{url}

\def\BibTeX{{\rm B\kern-.05em{\sc i\kern-.025em b}\kern-.08em
    T\kern-.1667em\lower.7ex\hbox{E}\kern-.125emX}}
\begin{document}

\title{Federated Action Recognition for Smart Worker Assistance Using FastPose\\
%\title{Federated Action Recognition using LSTM and Visual Transformers for Smart Worker Assistance}
\thanks{This work was funded by the Carl Zeiss Stiftung, Germany under the Sustainable Embedded AI project (P2021-02-009).}
}

\author{\IEEEauthorblockN{Vinit Hegiste}
\IEEEauthorblockA{\textit{Chair of Machine Tools and Control Systems} \\
\textit{RPTU Kaiserslautern-Landau}\\
Kaiserslautern, Germany \\
0000-0001-6944-1988}
\and
\IEEEauthorblockN{Vidit Goyal}
\IEEEauthorblockA{\textit{Chair of Machine Tools and Control Systems} \\
\textit{RPTU Kaiserslautern-Landau}\\
Kaiserslautern, Germany \\
0009-0007-8447-4666}
\and
\IEEEauthorblockN{Tatjana Legler}
\IEEEauthorblockA{\textit{Chair of Machine Tools and Control Systems} \\
\textit{RPTU Kaiserslautern-Landau}\\
Kaiserslautern, Germany \\
0000-0002-7297-0845}
\and
\IEEEauthorblockN{Martin Ruskowski}
\IEEEauthorblockA{\textit{Innovative Factory Systems (IFS)} \\
\textit{German Research Center for Artificial Intelligence (DFKI)}\\
Kaiserslautern, Germany \\
0000-0002-6534-9057}
}

\maketitle

\begin{abstract}
In smart manufacturing environments, accurate and real-time recognition of worker actions is essential for productivity, safety, and human–machine collaboration. While skeleton-based human activity recognition (HAR) offers robustness to lighting, viewpoint, and background variations, most existing approaches rely on centralized datasets, which are impractical in privacy-sensitive industrial scenarios. This paper presents a federated learning (FL) framework for pose-based HAR using a custom skeletal dataset of eight industrially relevant upper-body gestures, captured from five participants and processed using a modified FastPose model. Two temporal backbones, an LSTM and a Transformer encoder, are trained and evaluated under four paradigms: centralized, local (per-client), FL with weighted federated averaging (FedAvg), and federated ensemble learning (FedEnsemble). On the global test set, the FL Transformer improves over centralized training by +12.4 percentage points, with FedEnsemble delivering a +16.3 percentage points gain. On an unseen external client, FL and FedEnsemble exceed centralized accuracy by +52.6 and +58.3 percentage points, respectively. These results demonstrate that FL not only preserves privacy but also substantially enhances cross-user generalization, establishing it as a practical solution for scalable, privacy-aware HAR in heterogeneous industrial settings.
\end{abstract}


\section{Introduction}

In smart industrial environments, real-time recognition of worker actions plays a crucial role in enhancing productivity, ensuring safety, and enabling intelligent assistance systems. Skeleton-based action recognition has proved to be a robust alternative to RGB-based methods due to its invariance to lighting conditions, viewpoint changes, and background clutter~\cite{shahroudy2016ntu, ren2020survey_skeleton}. Using skeletal keypoints for human activity recognition (HAR) is well established, reducing input dimensionality and focusing on motion dynamics. Recent models often extract 2D pose sequences via OpenPose~\cite{cao2018pose}, YOLOv8-Pose~\cite{yolov8pose}, FastPose~\cite{zhang2019fastpose}, or BlazePose~\cite{bazarevsky2020blazepose}, and feed them into temporal architectures such as long short-term memory (LSTM) networks and vision transformers (ViTs)~\cite{plizzari2021spatial, ren2020survey_skeleton}.

LSTM networks have shown strong performance in modeling temporal dependencies in joint sequences~\cite{du2015hierarchical, liu2017gca_lstm}, while transformer-based architectures have gained popularity for their ability to model long-range interactions without recurrence~\cite{plizzari2021spatial}. However, training such models typically assumes access to large, centralized datasets. In industrial settings, action data is often collected from multiple sites or workers, each with potentially sensitive information. Centralizing such data not only poses logistical challenges but also raises significant privacy concerns, especially in safety-critical applications.

Federated learning (FL) addresses these challenges by enabling decentralized model training without transferring raw data~\cite{kairouz2021advances}. Instead, clients train local models on their private data and share only model updates with a central aggregator. This paradigm is particularly suitable for smart worker assistance systems, where data is inherently distributed and heterogeneous, and privacy-preserving solutions are crucial for deployment~\cite{hetrogenityfl2025legler}.  
Although several studies have explored FL for action recognition, most rely on simulated environments and public datasets. For example, Guo et al.~\cite{guo2023fsar} introduced FSAR, which combines adaptive topology learning with knowledge distillation to enhance FL performance on skeleton-based action recognition. Tu et al.~\cite{tu2024fedfslar} proposed FedFSLAR, focusing on few-shot federated action recognition using 3D-CNNs. However, these approaches neither utilize custom industrial datasets nor simulate realistic client distributions representative of worker-specific variability in manufacturing environments.

\begin{figure*}[htbp]
    \centering
    \includegraphics[width=0.8\textwidth]{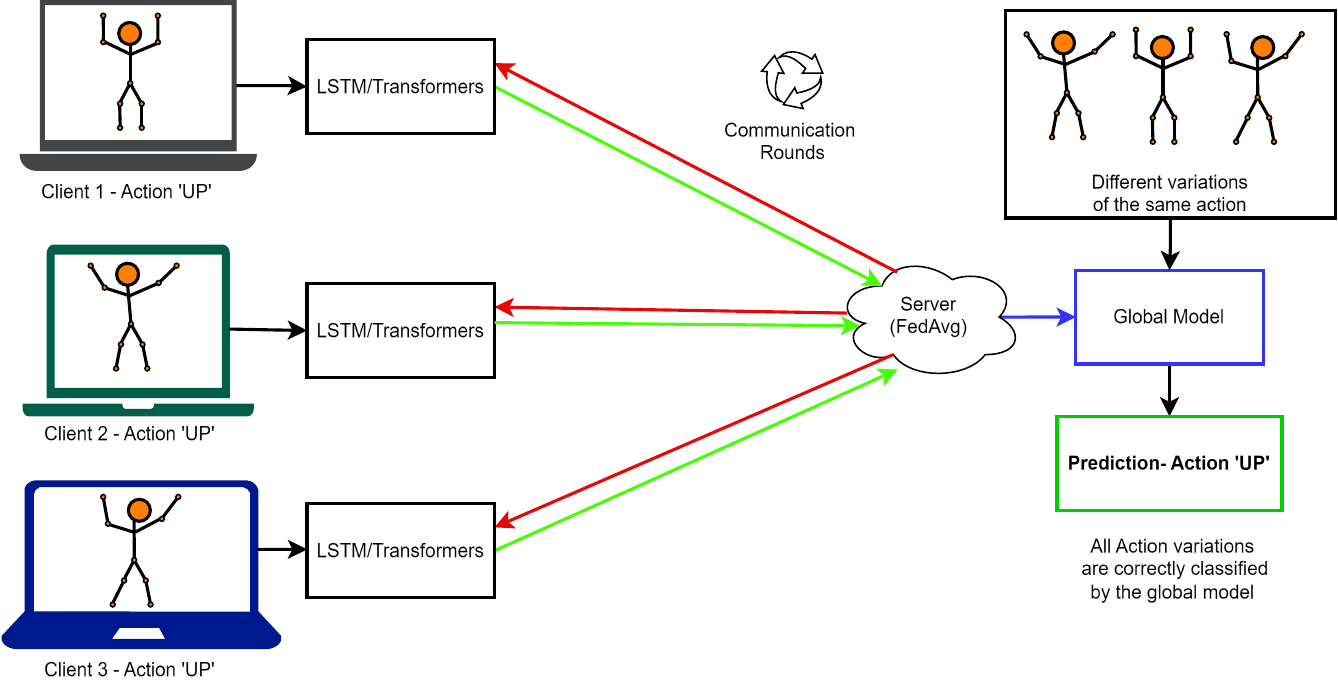}
    \caption{Overview of federated learning for skeleton-based human action recognition in smart manufacturing environments.}
    \label{fig:fedhar}
\end{figure*}

This work addresses these gaps by introducing a custom skeleton-based dataset of eight industrially relevant upper-body gestures, captured from five participants. Pose data is extracted using a modified FastPose model, and models are trained using both LSTM and vision transformer architectures over sliding windows of 20 frames. Each participant’s data is treated as a distinct client dataset, enabling realistic FL evaluation under client heterogeneity.  

\textbf{The key contributions of this paper are as follows:}
\begin{enumerate}
    \item We present a novel federated action recognition use case using a custom skeletal dataset tailored for smart worker assistance in industrial settings.
    \item We implement and evaluate two temporal models, LSTM and vision transformers under centralized, client local training, FL and FedEnsemble learning paradigms.
    \item We assess all models on a unified global test set compiled from held-out data of all participants, enabling a fair evaluation of generalization performance under client data heterogeneity.
    \item We also test all models on an unseen external participant to evaluate their generalization ability under domain shift, providing practical insight into deployment feasibility.
\end{enumerate}

Our objective is to develop a federated global model that generalizes across all participating clients and retains strong performance on heterogeneous, client-specific data, all while respecting privacy constraints and realistic industrial deployment conditions.

\section{Related Work}

Skeleton-based human action recognition (HAR) has emerged as a robust alternative to RGB-based approaches, offering compact and interpretable pose representations that are resilient to variations in lighting, background, and viewpoint~\cite{shahroudy2016ntu, ren2020survey_skeleton}. This modality is particularly valuable in privacy-sensitive domains such as smart manufacturing, where visual data cannot be easily shared. HAR pipelines typically begin with pose estimation, extracting 2D or 3D joint coordinates from video frames. Classical methods like OpenPose~\cite{cao2018pose} introduced part affinity fields for multi-person keypoint detection, while later models such as FastPose~\cite{zhang2019fastpose} and BlazePose~\cite{bazarevsky2020blazepose} improved efficiency for edge deployment. Recent solutions like YOLOv8-Pose~\cite{yolov8pose} unify object detection and pose estimation for real-time applications.

\subsection{Skeleton-Based Action Recognition}
\label{sec:skeleton_ar}

Earlier approaches to skeleton-based HAR relied on handcrafted features such as joint angles and motion trajectories~\cite{ren2020survey_skeleton}, but these lacked robustness across subjects and viewpoints. Deep learning methods, particularly recurrent neural networks (RNNs) and long short-term memory (LSTM) models, advanced the field by modeling temporal dynamics directly from raw keypoints. Du et al.~\cite{du2015hierarchical} proposed a hierarchical RNN to learn part-wise motion patterns, while Liu et al.~\cite{liu2017gca_lstm} introduced GCA-LSTM with attention mechanisms to emphasize salient joints. Graph convolutional networks (GCNs) further improved performance by modeling the human skeleton as a spatio-temporal graph, enabling structured reasoning across joints. Although not the focus of this work, GCNs remain a popular alternative to sequential models.

Recently, transformer-based models have gained traction due to their ability to capture long-range dependencies in a parallelized manner. Plizzari et al.~\cite{plizzari2021spatial} introduced a spatio-temporal transformer for joint and motion encoding, Aksan et al.~\cite{aksan2021slrformer} developed SLRFormer for sign language recognition, and Mehmood et al.~\cite{mehmood2024har_spatial} proposed ST-RTR to model relative joint movements. These architectures show promise for industrial HAR due to their scalability and expressive power.

\subsection{Federated Learning for Action Recognition}
\label{sec:fed_ar}
FL enables collaborative model training across distributed data silos without sharing raw data, addressing privacy and compliance concerns in industrial settings~\cite{kairouz2021advances}. FL has seen success in image classification and object detection for quality inspection~\cite{Hegiste2022applicationFL, Hegiste2023fedod, synthetic_fl}, as well as in language modeling~\cite{hard2019federatedlearningmobilekeyboard, Wen2023FLSurvey}. However, its application to HAR, particularly using skeletal data, remains underexplored.

Most FL-HAR studies focus on sensor or RGB modalities~\cite{2022Fedclar}, overlooking skeletal representations that offer both compactness and privacy advantages~\cite{ren2020survey_skeleton}. Furthermore, many rely on synthetic client splits from public datasets without considering real-world factors such as user heterogeneity, domain shifts, or unseen subjects. Guo et al.~\cite{guo2023fsar} proposed FSAR, an early FL framework for skeleton-based HAR that leverages adaptive graph topology and multi-grain knowledge distillation, but did not assess generalization to new users. Tu et al.~\cite{tu2024fedfslar} introduced FedFSLAR for few-shot video-based HAR using 3D CNNs on RGB inputs, without exploiting the advantages of skeletal data. FedCLAR~\cite{2022Fedclar} explored personalization via clustering for sensor data, yet relied heavily on public datasets such as \textit{NTU RGB+D} and \textit{OPPORTUNITY}, limiting practical transferability.

Existing FL-HAR approaches commonly assume idealized client distributions, lack evaluation under external test conditions, and rarely incorporate modern temporal backbones like transformers. To address these gaps, we develop a custom dataset simulating worker actions in smart manufacturing. Pose keypoints are extracted using FastPose, and clients are partitioned by subject to model realistic non-IID distributions. We evaluate both LSTM and transformer-based models under centralized, local, and federated settings, and assess generalization on an unseen user. This work contributes a realistic experimental framework and benchmark for federated skeleton-based HAR in industrial environments, highlighting the interaction between model architecture, data distribution, and deployment robustness.

\section{Methodology} \label{methodology}

The objective of this work is to evaluate FL-HAR using a custom dataset collected in an industrial context. Each participant is treated as an independent client in the FL setup, enabling training of a global model without sharing raw data. The performance of this global model is compared against locally trained per-client models and a centralized model trained on pooled data. This comparison demonstrates the benefits of collaborative learning over isolated training and serves as a motivator for potential industrial partners. Additionally, we assess model robustness using data from an external participant not involved in training.

\subsection{Dataset and Preprocessing}
\begin{figure}[htbp]
    \centering
    \includegraphics[width=0.5\textwidth]{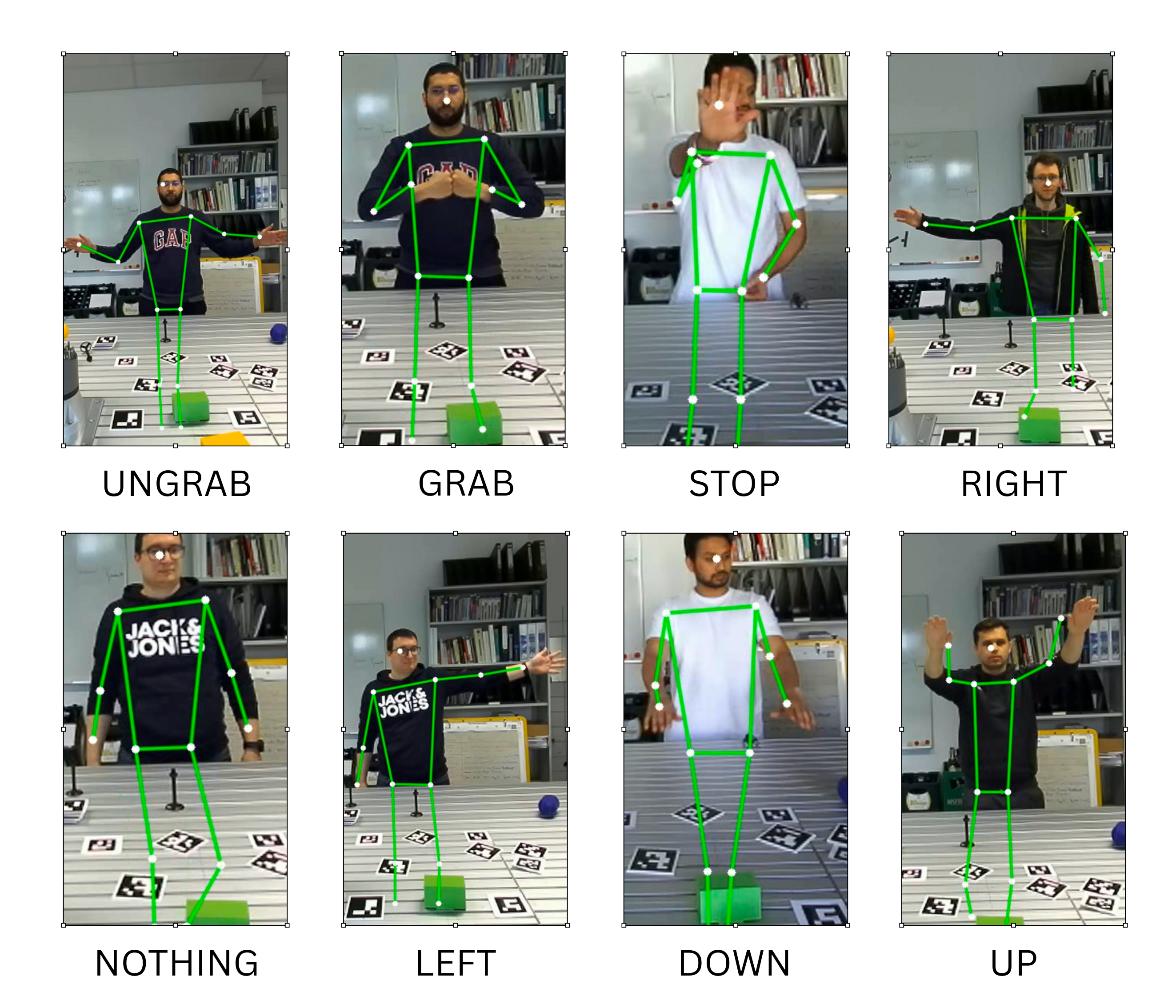}
    \caption{Snapshot of 8 Actions with FastPose Skeletal Points}
    \label{fig:gesture}
\end{figure}

We collected a custom pose-based HAR dataset comprising eight upper-body gestures: \emph{down}, \emph{grab}, \emph{left}, \emph{nothing}, \emph{right}, \emph{stop}, \emph{ungrab}, and \emph{up} (as shown in Fig.\ref{fig:gesture}) . Recordings were performed with five volunteer participants, each acting as a separate client in the FL setup. RGB-D video data was processed using a modified FastPose~\cite{zhang2019fastpose} pose estimation model, selected for its open-source availability, compact VGG-based architecture, and real-time CPU performance. FastPose is approximately 46\% smaller and 47\% faster than OpenPose~\cite{cao2018pose}, exceeding 50 FPS on a 2.7\,GHz Core i5 CPU, and sustaining 29–30 FPS in our deployment environment.
\begin{figure}[htbp]
    \centering
    \includegraphics[width=0.3\textwidth]{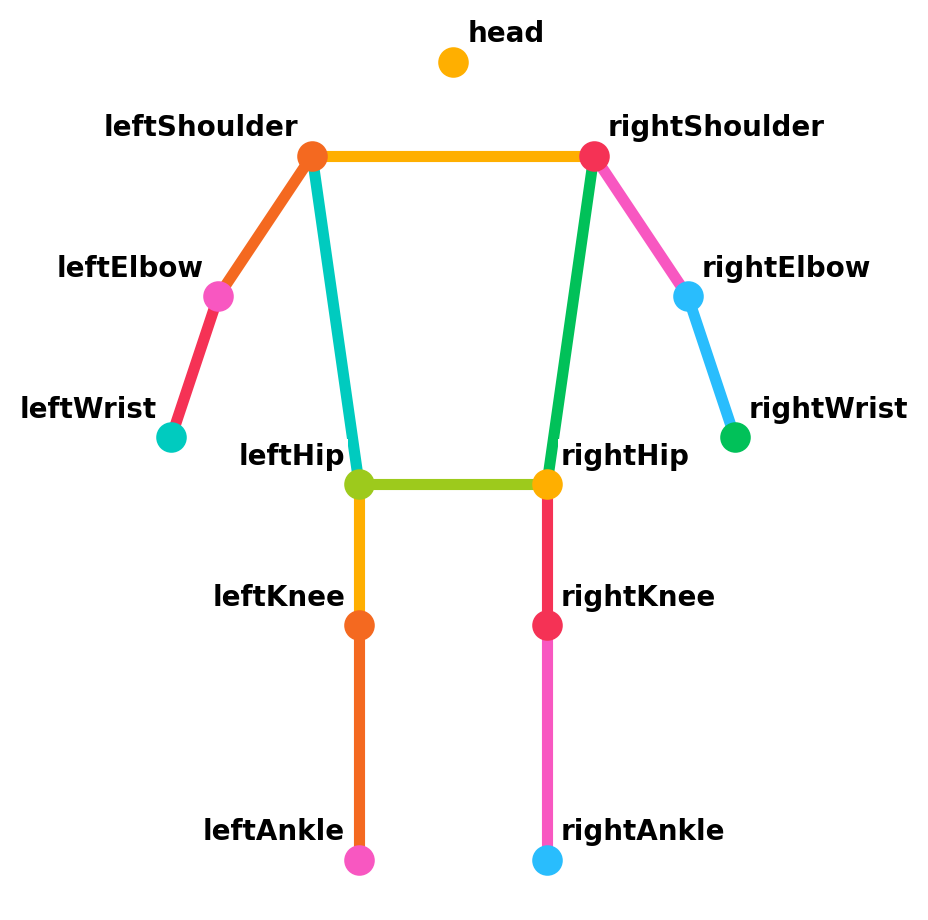}
    \caption{13 skeletal points used for Action Recognition}
    \label{fig:skeletal}
\end{figure}

In the original COCO-trained pipeline, 17 keypoints are estimated, including separate left/right eyes and ears~\cite{lin2014microsoft}. We simplify this by merging the four facial landmarks into a single \emph{head} point, yielding 13 joints in total (nose, head, shoulders, elbows, wrists, hips, knees, ankles) as shown in Fig. \ref{fig:skeletal}. This reduces input dimensionality, mitigates noise from occluded keypoints, and improves downstream processing efficiency without sacrificing recognition accuracy. Each frame is represented as a 26-dimensional vector (13 joints $\times$ 2 coordinates). %make sure to say this in presentation
\begin{figure}[htbp]
    \centering
    \includegraphics[width=\linewidth]{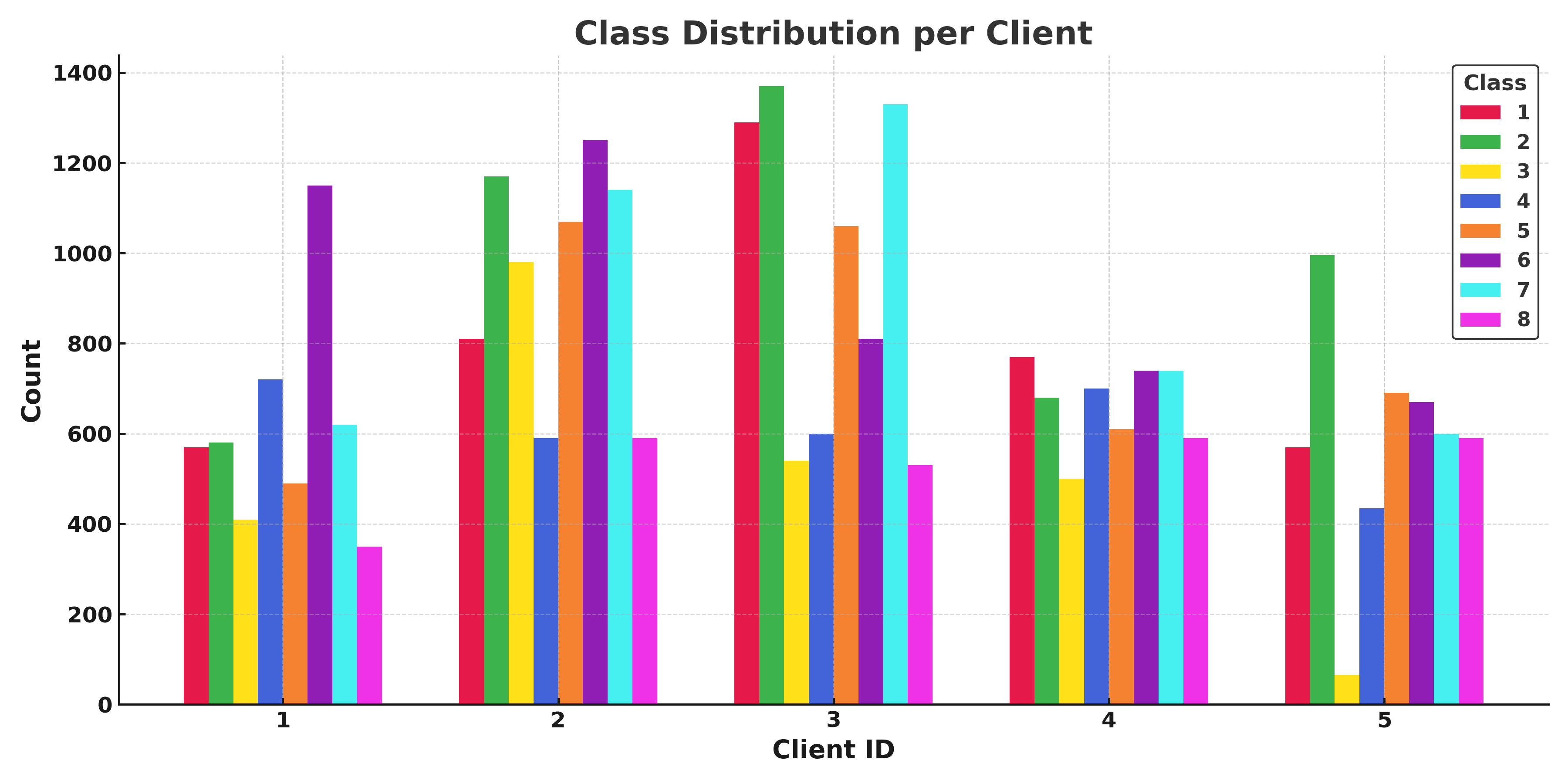}
    \caption{Class Distribution per Client.}
    \label{fig:class-client}
\end{figure}

\begin{figure*}[htbp]
    \centering
    \includegraphics[width=\linewidth]{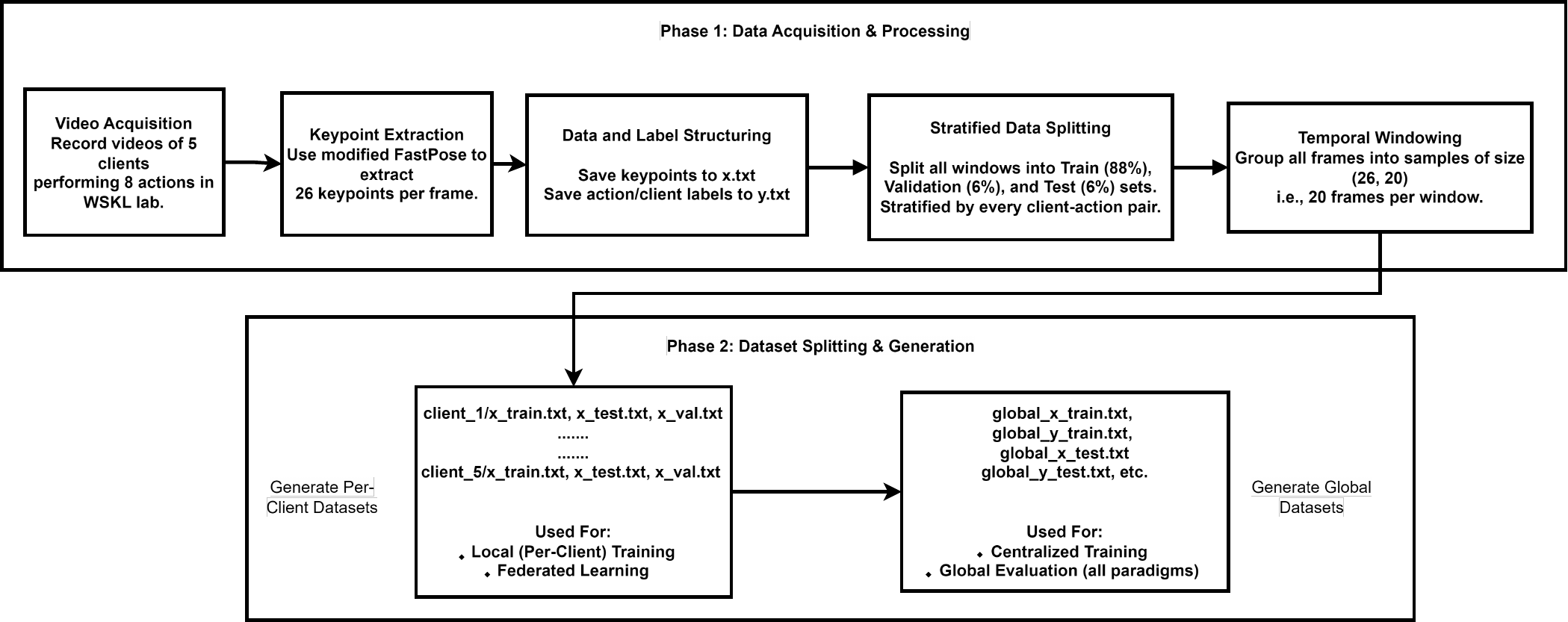}
    \caption{Federated action recognition pipeline with five clients. Each client trains locally on segmented pose sequences before model aggregation at the server.}
    \label{fig:dataflow}
\end{figure*}

The dataset contains 29{,}963 annotated frames across all gestures and clients. For each client, frames are stratified by gesture and split into training (88\%), validation (6\%), and test (6\%) subsets. Temporal dynamics are captured using a fixed-length sliding window of 20 consecutive frames, producing samples of shape $26 \times 20$. Each sample inherits the action label and the client ID. On average, each client has $\sim$6{,}000 samples, though counts vary with gesture frequency and recording duration. This setup produces a realistic non-IID distribution, with differences in gesture proportions and motion styles between clients and can be referred in Fig.\ref{fig:class-client}. Both per-client datasets (for FL and local training) and a pooled dataset (for centralized training) are prepared. A summary of the workflow is shown in Fig.~\ref{fig:dataflow}. The final processed dataset of each client are also uploaded to hugging face for public access\footnote{https://huggingface.co/datasets/WSKL/FederatedHAR/tree/main}. %use WSKL hugging face

\subsection{Model Architectures and Training} \label{model_arch}
We designed two simple and small temporal sequence models for classifying the 20-frame pose windows: a recurrent LSTM and a Transformer encoder. The LSTM consists of two stacked recurrent layers (hidden size 128) followed by a fully connected softmax output layer. The Transformer comprises four self-attention encoder layers (model dimension 128, four heads), operating on the sequence of pose vectors after positional encoding. Both models are implemented in PyTorch Lightning, trained with the Adam optimizer (learning rate $2\times10^{-4}$, batch size 64), and use standard weight initialization. No pre-training or data augmentation beyond pose estimation was applied. Models are trained on CPU-only hardware to simulate edge deployment constraints, with early stopping (patience 15 epochs) and check pointing enabled. The idea behind this was to have more emphasis on the FL part than the model architecture it self.

\subsection{Federated Learning Setup}
In the FL setup, each participant corresponds to one client, resulting in a five-client configuration. At each communication round $t$, clients train locally for a fixed number of epochs and send updated model weights $w_i^{(t)}$ to the server. The server aggregates these using the weighted Federated Averaging (FedAvg) rule:
\begin{equation}
w^{(t+1)} = \sum_{i=1}^{K} \frac{n_i}{N}\,w_i^{(t)},
\label{eq:fedavg}
\end{equation}
where $n_i$ is the number of samples at client $i$ and $N=\sum_{i=1}^K n_i$. This ensures that clients with more data have proportionally greater influence on the global update. Both the FL setup and the FedEnsemble models use this aggregation rule, differing only in the initial data partitioning across clients. All the 5 clients participate in each communication round due to low number of clients.

\subsection{Experimental Setup}
To evaluate the impact of different training strategies on model generalization, we compared four paradigms: FL, local (per-client) training, centralized learning, and FedEnsemble learning using a unified global test set. The global test set was compiled by combining the held-out test data from all five clients, ensuring identical evaluation conditions across paradigms. Each setup was designed to maintain a comparable computational budget, where the total number of training epochs is calculated as:
\[
\text{Total rounds} = \text{Local epochs} \times \text{Communication rounds}.
\]
This ensured that differences in performance could be attributed solely to the training strategy rather than unequal training time or parameter tuning.

\subsubsection{Federated Learning}
In the standard FL setup, each of the five clients trained locally on its own dataset for 25 epochs per communication round. After each round, the locally updated model weights were sent to the central server, where they were aggregated using the weighted FedAvg rule from Eq.~\eqref{eq:fedavg}. This process was repeated for 20 communication rounds, giving an equivalent of \(20 \times 25 = 500\) total local epochs per client, matching the training budget of the centralized baseline. Only model weights and the number of samples per client were shared; no raw data or class distribution statistics were exchanged, preserving data privacy. The final global model obtained after the last communication round was evaluated on the unified global test set.

\subsubsection{Local (Per-Client) Training}
For the local baseline, separate LSTM and Transformer models were trained independently on each client’s training and validation sets for up to 500 epochs, using early stopping with a patience of 15 epochs to prevent overfitting as the clients on their own had much smaller dataset. This produced five distinct models, one per client without any collaboration or parameter sharing. Each model was then evaluated on all five client test sets and for generalizability the models were also tested with the global test dataset.

\subsubsection{Centralized Training}
In the centralized baseline, the training data from all five clients was pooled into a single dataset, removing any client-specific data separation. Both LSTM and Transformer architectures were trained on this combined dataset for 500 epochs, ensuring the same total training budget as in FL. The trained model was then evaluated on the unified global test set. This scenario represents the idealized case without privacy constraints, where all training data is available to a single learning process.

\subsubsection{Federated Ensemble Learning}
The FedEnsemble configuration was inspired by prior ensemble-based FL approaches~\cite{fedesembleyolo,synthetic_fl}. In this setup, the centralized dataset was uniformly partitioned into five client datasets of equal size, each containing a mix of data from at least three original participants. This created an IID distribution across clients while ensuring identical overall data coverage to the centralized baseline. Training was performed using the FedAvg protocol for 20 communication rounds with 25 local epochs per round, mirroring the FL setup. Since privacy is not a factor in this configuration, the purpose of this experiment was to isolate and assess the benefits of FL as an ensemble learning method, particularly for small datasets as often encountered in manufacturing.

All experiments used identical model architectures, optimizer configuration, batch sizes, and total training budgets (\(500\) total rounds) as mentioned in \ref{model_arch}. No data augmentation was applied beyond the preprocessing provided by the pose estimation pipeline. All models were trained and evaluated on CPU-only hardware to simulate deployment in resource-constrained edge environments.

\section{Results and Discussion}
\begin{table*}[htbp]
  \centering
  \footnotesize % Slightly larger than \scriptsize
  \setlength{\tabcolsep}{5pt} % Moderate padding
  \caption{Comparative Accuracy of Learning Paradigms on the Global Test Set}
  \label{tab:comparative_accuracy}
  \begin{adjustbox}{width=0.8\textwidth,center}
    \begin{tabular}{@{} llcc @{}}
      \toprule
      \textbf{Training Paradigm} 
        & \textbf{Model Architecture} 
        & \makecell{\textbf{Global Test}\\\textbf{Accuracy (\%)}} 
        & \makecell{\textbf{Performance}\\\textbf{Delta vs.\ Centralized}} \\
      \midrule
      \multirow{2}{*}{Local (Per‑Client)} 
        & LSTM        & $\sim50\%$ (own), $<20\%$ (other) & N/A \\
        & Transformer & $\sim50\%$ (own), $<20\%$ (other) & N/A \\
      \addlinespace
      \multirow{2}{*}{Centralized} 
        & LSTM        & 50.0                               & Baseline \\
        & Transformer & 57.1                               & Baseline \\
      \addlinespace
      \multirow{2}{*}{Federated (FedAvg)} 
        & LSTM        & 59.9                               & +9.9     \\
        & Transformer & 69.5                               & +12.4    \\
      \addlinespace
      \multirow{2}{*}{Ensemble Federated} 
        & LSTM        & 61.6                               & +11.6    \\
        & Transformer & 73.4                               & +16.3    \\
      \bottomrule
    \end{tabular}
  \end{adjustbox}
\end{table*}
Global test accuracy of LSTM vs.\ Transformer models under different training paradigms is summarized in Table \ref{tab:comparative_accuracy}. The results of all algorithms on global test dataset are explain in more details in further subsections.

\subsection{Federated Learning}
Remarkably, FL outperformed centralized training despite never sharing raw data. The federated Transformer attained 69.5\% test accuracy (+12.4\%pp over centralized), and the federated LSTM reached 59.9\% (+9.9\%). We attribute this boost to a regularization‐like effect: aggregating diverse local updates prevents overfitting to any single client’s bias. By structuring training as 20 rounds of 25 local epochs (totaling 500 total rounds), FedAvg matched the centralized training budget while harnessing complementary knowledge from each client. This demonstrates FL’s potential to leverage distributed heterogeneity for improved generalization.

\subsection{Local (Per‑Client) Training}
Each client’s model achieved below par accuracy on its own held‑out data, and also failed to generalize to other users. Off‑diagonal entries in the \(5\times5\) accuracy matrix frequently dropped below 20\%, with some as low as 0\% (e.g., a model trained on Client 1 scored 0\% on Client2 Test set). This severe degradation highlights overfitting to personal motion patterns and confirms that collaborative training is necessary for robust cross‑user HAR models. The Transformer slightly outperformed the LSTM locally, but both suffered from poor cross‑client transfer, indicating that rather than model capacity, insufficient data was the root issue.

\begin{figure}[htbp]
    \centering
    \includegraphics[width=0.45\textwidth]{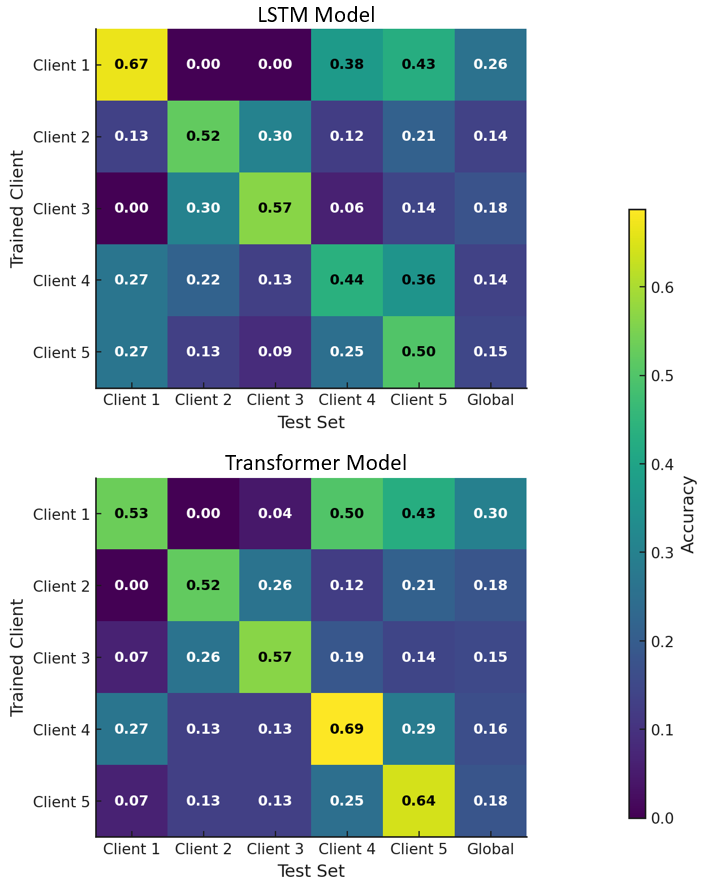}
    \caption{Results of Cross-Client Validation with Local and Global Test set}
    \label{fig:local}
\end{figure}

\subsection{Centralized Training}
Training on the pooled dataset yielded better generalization: the centralized Transformer achieved 57.1\% accuracy on the global test set, while the centralized LSTM reached around 50\%. The Transformer’s superior performance aligns with its ability to capture global temporal patterns more effectively than recurrent networks. Nevertheless, 57.1\% leaves substantial room for improvement, suggesting that simple pooling did not fully resolve data diversity issues. Underrepresented clients may have been dominated by more frequent patterns, leading to biased predictions.

\subsection{Federated Ensemble Learning}
In the FedEnsemble scenario, the federated Transformer achieved 73.4\% accuracy and the federated LSTM 61.6\%. This setup used the exact same training dataset as centralized learning, which was uniformly partitioned among clients, meaning privacy preservation was not a factor. The results show that applying FL as an ensemble learning method yields higher accuracy than centralized learning, an improvement of +16.3\% for the Transformer and +11.6\% for the LSTM over their centralized counterparts. This indicates that the ensemble effect of bagging and boosting by combining multiple independently trained client models, which can extract more robust representations than a single centralized model, particularly when working with small datasets, which is common in manufacturing scenarios. Such a strategy could be adopted as a standard practice when data volume is limited.

\subsection{External Client Evaluation}
\begin{figure*}[htbp]
    \centering
    \includegraphics[width=\textwidth]{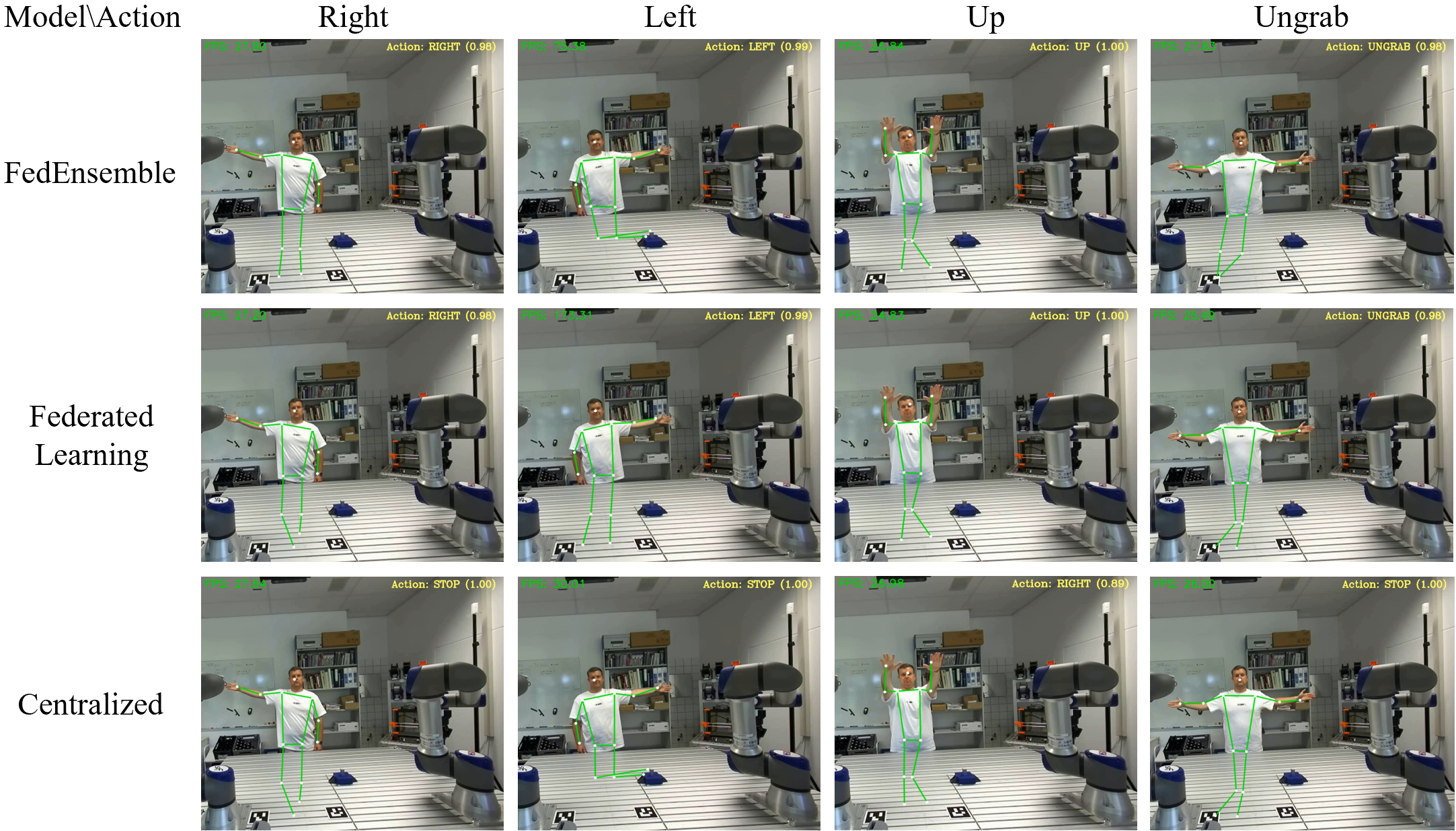}
    \caption{Results of FL, FedEnsemble and Centralized Transformer on Unseen External Client Actions}
    \label{fig:fed_external_client}
\end{figure*}

To assess real-world generalization, all centralized, FL and FedEnsemble with Transformer models were evaluated on an external client test set corresponding to an unseen subject, whose data was excluded from training in all paradigms. Here we only selected Transformer architecture models as it performed better compared to the LSTM models. As shown in Table~\ref{tab:external_client}, both federated approaches substantially outperformed centralized training. The FL Transformer achieved 64.29\% accuracy, a +52.58\% point gain over its centralized counterpart, while the FedEnsemble Transformer further improved to 69.98\%, a +58.27\% point gain. The centralized model just predicted `Stop' label for most of the actions, resulting in very poor performance.
The snapshots from the results from few of actions as shown in Fig. \ref{fig:fed_external_client}. It can be seen that both FedEnsemble and FL models perform well and also predict the actions with high confidence score.
This demonstrates that FL not only preserves privacy but also yields models that transfer better to new users, likely due to the diversity of local updates acting as a regularizer. 
\begin{table}[htbp]
  \centering
  \footnotesize
  \caption{Performance on External Client Test Set (Unseen Subject) for Transformer}
  \label{tab:external_client}
  \begin{adjustbox}{width=0.3\textwidth,center}
    \begin{tabular}{@{} lc @{}}
      \toprule
      \textbf{Training Paradigm} & \textbf{Accuracy} \\
      \midrule
      FedEnsemble Learning  & 69.98\% \\
      Federated Learning  & 64.29\% \\
      Centralized         & 11.71\% \\              
      \bottomrule
    \end{tabular}
  \end{adjustbox}
\end{table}

\subsection{Summary of Results}
Overall, the experiments confirm that FL is highly effective for pose-based HAR in heterogeneous industrial settings, consistently outperforming both centralized training and local models. On the global test set, FL with Transformer achieved a +12.4\% gain over centralized learning, while FedEnsemble pushed this to +16.3\%, demonstrating the added benefit of combining independently trained models. The external client evaluation further validated these trends, with FL improving accuracy by over 50 percentage points and FedEnsemble by nearly 60 percentage points compared to centralized training. These gains highlight FL’s dual advantage: preserving privacy while producing models that generalize better to unseen users. The results also establish FedEnsemble as a viable strategy for boosting performance even in non-privacy scenarios, making it especially valuable when data is scarce or distributed across isolated sources. Together, these findings position FL and FedEnsemble as strong candidates for scalable, privacy-aware HAR solutions in real-world manufacturing environments.

\section{Conclusion}
This paper investigated FL for pose-based HAR in industrial contexts, introducing a custom dataset of eight upper-body gestures recorded from five participants, each acting as an independent client. The pipeline combined RGB-D video capture, 2D skeletal keypoint extraction via a modified FastPose with a compact 13-joint representation, and segmentation into 20-frame windows to emulate realistic non-IID client data. Two temporal architectures, LSTM and Transformer encoders, were trained under four paradigms: centralized learning, local per-client training, FL with weighted FedAvg, and FedEnsemble learning. Models were evaluated on a unified global test set as well as an unseen external participant to measure generalization under domain shift.

Across all evaluations, FL consistently outperformed centralized learning while preserving data privacy. On the global test set, the FL Transformer achieved 69.5\% accuracy (+12.4\% over centralized), and the FL LSTM reached 59.9\% (+9.9\%). FedEnsemble learning, using the exact same dataset as centralized training but uniformly partitioned into IID client splits, further improved results to 73.4\% (Transformer) and 61.6\% (LSTM), highlighting the benefit of model aggregation as an ensemble mechanism even when privacy is not a factor.

External client testing reinforced these trends: the FL Transformer reached 64.29\% accuracy and the FedEnsemble Transformer achieved 69.98\%, compared to only 11.71\% for centralized training. This substantial improvement demonstrates that FL not only supports privacy-preserving collaboration but also yields models that transfer better to unseen users, likely due to the diversity of local updates acting as a regularizer.

Overall, the results establish FL as both a privacy-preserving and performance-enhancing paradigm for cross-user HAR in heterogeneous industrial environments. The observed ensemble effect in FedEnsemble suggests that, even without privacy constraints, FL can be exploited as a robust training strategy for small, distributed datasets common in manufacturing. Future work will scale this framework to larger client populations, incorporate advanced aggregation and personalization methods, and integrate multi-sensor fusion to further improve robustness and real-world deployability.

\section{Limitations and Outlook}
The present study has several limitations that provide avenues for future research. First, the dataset size constrained the number of simulated clients to five, limiting both the diversity of motion patterns and the degree of domain heterogeneity that can be modeled. Expanding the dataset to include between 10 and 15 clients with varying anthropometric characteristics, gesture speeds, and environmental conditions would enable a more representative evaluation of real-world deployments. Second, only the weighted FedAvg aggregation strategy was employed. While it demonstrated strong performance, exploring advanced federated optimization methods such as FedProx, FedAdam, FedBN, and personalized FL approaches may further improve robustness under non-IID and imbalanced client data distributions. Third, the study relied exclusively on RGB-D based skeletal pose estimation, which may be affected by occlusion, lighting variations, or sensor noise. Incorporating multi-sensor fusion (e.g., IMUs or depth-only processing) could enhance robustness in challenging industrial environments.

Building on these findings, our ongoing research focuses on deploying the federated pose-based HAR system in a robotics application. Specifically, we are integrating the trained models into a ROS~2 pipeline to control a robotic arm based on recognized human gestures. The system will stream RGB-D camera data to perform real-time pose estimation via the modified FastPose, followed by on-device classification, and will translate predicted actions (e.g., \emph{grab}, \emph{left}, \emph{stop}, \emph{up}) into corresponding arm control commands. This deployment targets assistive technology scenarios where the robot responds promptly to user gestures, ensuring low-latency inference and robust closed-loop performance. ROS~2 will be leveraged for efficient sensor data handling, real-time processing, and seamless integration between the machine learning module and the robot motion controllers.

\textbf{Outlook:} Beyond robotics, the proposed FL-HAR framework can be extended to collaborative manufacturing cells, remote training environments, and safety monitoring systems, where privacy-preserving, real-time human action recognition is critical. Integrating adaptive client weighting, federated domain adaptation, and lightweight on-device inference models will be key to enabling large-scale, heterogeneous deployments. In the long term, this line of research can contribute to establishing standardized FL benchmarks for industrial HAR, enabling reproducible evaluation and accelerating adoption across industries.

%From a system perspective, all training and evaluation were conducted under controlled laboratory conditions, and the real-time constraints of field deployment have yet to be validated. The current models were trained on static client partitions; in practical settings, client membership may change over time (e.g., new workers joining), requiring adaptive or continual FL strategies. Additionally, the evaluation focused primarily on accuracy; future work should also assess latency, model size, energy consumption, and fault tolerance in edge computing scenarios.

\section*{Acknowledgment}
This work was funded by the Carl Zeiss Stiftung, Germany under the Sustainable Embedded AI project (P2021-02-009).
%The authors would like to sincerely thank the volunteer participants for their time and effort in performing the actions used to create the custom skeletal dataset for this study. Their cooperation made the development and evaluation of the proposed FL-HAR framework possible.

\bibliographystyle{IEEEtran}
\bibliography{lit}

\end{document}